\documentclass[letterpaper, 10 pt, conference]{ieeeconf}

\IEEEoverridecommandlockouts

\usepackage{cite}
\usepackage{amsmath,amssymb,amsfonts}
\usepackage{algorithmic}
\usepackage{graphicx}
\usepackage{textcomp}
\usepackage{xcolor}
\usepackage{siunitx}
\usepackage{float}
\usepackage[symbol]{footmisc}
\usepackage{ar}
\usepackage{balance}
\usepackage{hyperref}
\hypersetup{
    colorlinks=true,
    linkcolor=black,
    filecolor=none,      
    urlcolor=blue,
    citecolor=black,
    pdftitle={Overleaf Example},
    pdfpagemode=FullScreen,
    }

\newcommand{\bs}[1]{\boldsymbol{#1}}  
\newcommand{\ts}[1]{\text{#1}} 

\renewcommand{\baselinestretch}{0.8236}

\def\BibTeX{{\rm B\kern-.05em{\sc i\kern-.025em b}\kern-.08em

    T\kern-.1667em\lower.7ex\hbox{E}\kern-.125emX}}

\begin{document}

\title{\LARGE \bf Progress Towards Submersible Microrobots: A Novel \mbox{$\bs{13}$-mg} \mbox{Low-Power} \mbox{SMA-Based} Actuator for Underwater Propulsion \\

\thanks{This work was partially funded by the Washington State University (WSU) Foundation and the Palouse Club through a Cougar Cage Award to \mbox{N.\,O.\,P\'erez-Arancibia}. Additional funding was provided by the WSU Voiland College of Engineering and Architecture through a start-up fund to \mbox{N.\,O.\,P\'erez-Arancibia}.} %
\thanks{C.\,R.\,Longwell and C.\,K.\,Trygstad contributed equally to this work.} %
\thanks{The authors are with the School of Mechanical and Materials Engineering, Washington State University (WSU), Pullman,\,WA\,99164,\,USA. Corresponding authors' \mbox{e-mail:}
{\tt conor.trygstad@wsu.edu}~(C.\,K.\,T.);
\\
{\tt n.perezarancibia@wsu.edu} (N.\,O.\,P.-A.).}%
}
\author{Cody R. Longwell, Conor K. Trygstad, Francisco M.\,F.\,R. Gon\c{c}alves, Ke Xu, and N\'estor O. P\'erez-Arancibia}

\maketitle
\thispagestyle{empty}
\pagestyle{empty}

\begin{abstract}
We introduce a new \mbox{low-power} \mbox{$\bs{13}$-mg} microactuator driven by \textit{\mbox{shape-memory} alloy} (SMA) wires for underwater operation. The development of this device was motivated by the recent creation of microswimmers such as the \mbox{FRISHBot}, \mbox{WaterStrider}, \mbox{VLEIBot}, \mbox{VLEIBot\textsuperscript{+}}, and \mbox{VLEIBot\textsuperscript{++}}. The first four of these robots, ranging from \mbox{$\bs{30}$} to \mbox{$\bs{90}$~mg}, function tethered to an electrical power supply while the last platform is an \mbox{$\bs{810}$-mg} fully autonomous system. These five robots are driven by \textit{dry} \mbox{SMA-based} microactuators first developed for microrobotic crawlers such as the \mbox{SMALLBug} and \mbox{SMARTI}. As shown in this abstract, dry \mbox{SMA-based} actuators do not operate efficiently under water due to high \mbox{heat-transfer} rates in this medium; for example, the actuators that drive the \mbox{VLEIBot\textsuperscript{++}} require about \mbox{$\boldsymbol{40}$\,mW} of average power at \mbox{$\boldsymbol{1}$\,Hz} in dry air while requiring about \mbox{$\boldsymbol{900}$\,mW} of average power at \mbox{$\boldsymbol{1}$\,Hz} in water. In contrast, the microactuator presented in this abstract consumes about \mbox{$\bs{150}$\,mW} of average power at \mbox{$\boldsymbol{1}$\,Hz} in both dry air and water; additionally, it can be excited directly using an onboard battery through simple power electronics implemented on a \mbox{custom-built} \textit{printed circuit board} (PCB). This technological breakthrough was enabled by the integration of a soft structure that encapsulates the SMA wires that drive the actuator in order to passively control the rates of heat transfer. The results presented here represent preliminary, yet compelling, experimental evidence that the proposed actuation approach will enable the development of fully autonomous and controllable submersible microswimmers. To accomplish this objective, we will evolve the current version of the \mbox{VLEIBot\textsuperscript{++}} and introduce new bioinspired underwater propulsion mechanisms.
\end{abstract}

\section{Introduction} 
\label{SECTION01}
\begin{figure}[t!]
\vspace{2ex}
\begin{center}
\includegraphics[width=0.48\textwidth]{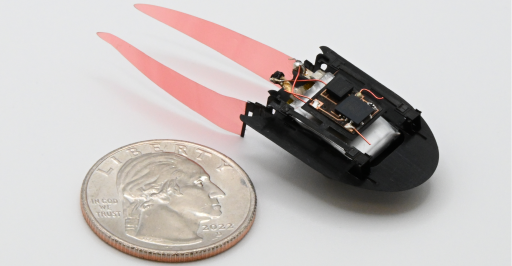}
\vspace{-3ex}
\caption{\textbf{Photograph of the VLEIBot\textsuperscript{++}.} This microrobot is an \mbox{$810$-mg} surface swimmer driven by a dry \mbox{SMA-based} actuator that can function autonomously from both the power and control perspectives (see accompanying supplementary movie). \label{FIG01}}
\end{center}
\vspace{-3ex}
\end{figure}

Submersible aquatic microrobots have significant potential to assist humans with complex tasks, including search and rescue, ecological monitoring, surveillance, underwater inspection of infrastructure and machinery, maintenance of aquaculture instrumentation and facilities, and automated pest control in hydroponic agriculture. For this vision to become a reality, the microswimmers must operate autonomously, have structural and behavioral robustness, and be resilient against environmental uncertainty. The recent development of novel \textit{\mbox{high-work-density}} (HWD) microactuators driven by \emph{\mbox{shape-memory} alloy} (SMA) wires enabled the creation of numerous locomoting microrobots with new \mbox{capabilities---for} example, see~\cite{RoBeetle_2020,SMALLBug_2020,SMARTI_2021,WaterStrider_2023,VLEIBot_2024,FRISHBot_2024,VLEIBot++_2024}. The newest of these robotic systems is the \mbox{VLEIBot\textsuperscript{++}} surface swimmer shown in Fig.\,\ref{FIG01}, which weighs \mbox{$810$\,mg}, is driven by \textit{dry} \mbox{SMA-based} actuators, and can function autonomously from both the power and control perspectives (see accompanying supplementary movie). Preliminary experiments indicate that due to high \mbox{heat-transfer} rates in water and associated power inefficiencies, the dry actuators that drive the \mbox{VLEIBot\textsuperscript{++}} are not well suited to propel submersible swimmers. To accurately understand these phenomena, we designed and performed experiments that quantify the power requirements for dry actuators to operate in air and water. Furthermore, we propose a new design for an actuator that can operate in air and water with approximately the same power requirements.
\begin{figure*}[t!]
\vspace{2ex}
\begin{center}
\includegraphics[width=0.96\textwidth]{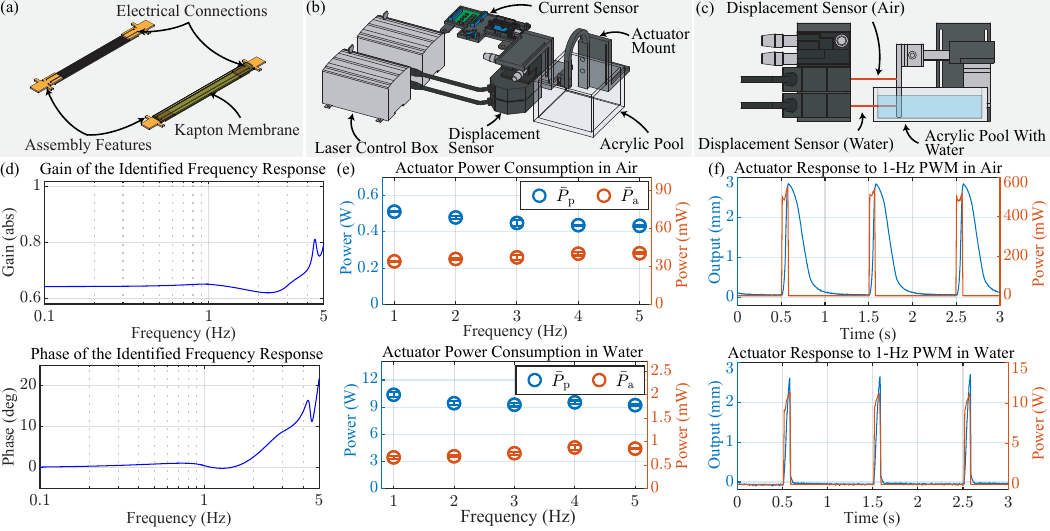}
\end{center}
\vspace{-3ex}
\caption{\textbf{Experimental measurement of the power consumed by \mbox{SMA-based} microactuators during operation in air and water.} \textbf{(a)}~Dry \mbox{SMA-based} microactuator developed to drive \mbox{VLEIBot-like} swimmers and characterized through the power experiments discussed in this abstract (top), and conceptual design of a \mbox{low-power} underwater \mbox{SMA-based} microactuator (bottom). \textbf{(b)}~Experimental setup used to measure displacement and power consumption of the tested \mbox{SMA-based} microactuator in both air and water. The output displacement of the actuator is measured using a Keyence LK\,$031$ laser sensor and the current used to compute power is measured using an Adafruit INA260 sensor. \textbf{(c)}~Hardware configuration used to collect the data for identifying the system that maps the true actuation output to the measurement distorted by the path of the sensing laser (water and acrylic). \textbf{(d)}~Identified model of the sensing system described in (c). \textbf{(e)}~Mean and \textit{experimental standard deviation} (ESD) of the average ($P_{\ts{a}}$) and peak ($P_{\ts{p}}$) power consumption of dry \mbox{SMA-based} actuator tested in air and water. \textbf{(f)}~Actuator response and power consumption in air and water for a \mbox{$1$-Hz} PWM excitation. \label{FIG02}}
\vspace{-3ex}
\end{figure*}

\section{Dry Actuation and \mbox{Power-Consumption} Characterization in Air and Water}
\label{SECTION02}
The dry microactuators that drive swimmers of the \mbox{VLEIBot} type, used in the research presented here and depicted at the top of~\mbox{Fig.\,\ref{FIG02}(a)}, are made from carbon fiber, CuFR$4$ material, and SMA \mbox{NiTi} wires with a diameter of \mbox{$38.1$\,{\textmu}m} and nominal transition temperature of $90\,^{\circ}\ts{C}$~\cite{VLEIBot_2024}. Given the thermal nature of \mbox{SMA-based} actuation, the corresponding energy and power requirements directly depend on \mbox{heat-transfer} phenomena. We quantified the power consumption of the tested actuators in air and water using the experimental setup shown in~\mbox{Fig.\,\ref{FIG02}(b)}, which includes sensors that simultaneously measure electrical current (Adafruit, INA260) and actuator displacement (Keyence, LK\,$031$). The actuators are excited using controlled \textit{\mbox{pulse-width} modulation} (PWM) signals generated with the Mathworks Simulink real-time host--target system described in~\cite{WaterStrider_2023,VLEIBot_2024}. This experimental setup was also used to quantify the power requirements of a new underwater HWD actuator with the basic design shown at the bottom of~\mbox{Fig.\,\ref{FIG02}(a)} and discussed in \mbox{Section\,\ref{SECTION03}}. In preliminary experiments, we observed that using a laser displacement sensor through an acrylic pool and water, as shown in~\mbox{Fig.\,\ref{FIG02}(c)}, distorts the obtained measurement. To understand and compensate for this effect, we estimated a dynamical model that maps the true signal into the distorted measurement by performing a \mbox{discrete-time} \textit{linear \mbox{time-invariant}} (LTI) system identification procedure based on \mbox{least-squares} minimization and assuming an \textit{\mbox{infinite-impulse response}} (IIR) structure (see Chapter\,5.1.2 in~\cite{Keesman2011SystemID}). \mbox{Fig.\,\ref{FIG02}(d)} shows the frequency response of an identified IIR model with order $100$. The input data used for system identification is frequency rich (persistently exciting) up to \mbox{$5$\,Hz}; therefore, we consider the Bode plot in~\mbox{Fig.\,\ref{FIG02}(d)} to be valid only at low frequencies (\mbox{$<5\,\ts{Hz}$}). For all practical purposes, at low frequency, the identified system is simply a static mapping with a gain of $0.633$. We further validated these findings using the linear regression function of the \mbox{scikit-learn} Python library to fit a \textit{\mbox{finite-impulse} response} (FIR) filter with order $256$.

In both air and water, we performed five types of \mbox{power-consumption} \mbox{$32$-second} experiments corresponding to the PWM \mbox{frequency--duty-cycle} pairs, \mbox{$\left\{f(i),\ts{DC}(i)\right\}$}, with the index \mbox{$i \in \left\{ 1, \cdots, 5\right\}$}, for \mbox{$ f \in \left[1\hspace{-0.3ex}:\hspace{-0.3ex}1\hspace{-0.3ex}:\hspace{-0.3ex}5\right]$\,Hz} and \mbox{$\ts{DC} \in \left[7,8,9,10,10\right]$\,\%}, while measuring and recording both the actuator output and electrical current, according to the scheme in~Fig.\,\ref{FIG02}(b). In air, we excited the actuator using a voltage---typically with a maximum value of \mbox{$2.7\,\ts{V}$}---that nominally generates a maximum current of \mbox{$250$\,mA} through its SMA wires. In water, heuristically through simple experiments, we set a voltage that maximizes actuation output but is sufficiently low---typically with a maximum value of \mbox{$12\,\ts{V}$}---such that the SMA wires of the device do not burn. After collecting the experimental data, we converted the currents into power signals according to \mbox{$P(t) = I^2(t)\cdot R$}, where respectively $I(t)$ and $R$ are the measured instantaneous current flowing through the SMA wires of the device and total constant resistance of the actuator's circuit (SMA + power wires), and $t$ denotes time. For each experiment, we computed the average consumed power, $P_{\ts{a}}$, and peak power, $P_{\ts{p}}$, using $30$ seconds of \mbox{steady-state} data. The estimated temperature of both media during all the experiments was on the order of $23\,^{\circ}\ts{C}$ and the measured actuation outputs exceeded $2\,\ts{mm}$ at $1\,\ts{Hz}$ (see \mbox{Fig.\,\ref{FIG02}(f)}).

For each tested PWM pair $\left\{f,\ts{DC}\right\}$, we performed five \mbox{back-to-back} experiments; then, we computed the mean and \textit{experimental standard deviation} (ESD) of both $P_{\ts{a}}$ and $P_{\ts{p}}$. These data are summarized in~\mbox{Fig.\,\ref{FIG02}(e)}, where 
$\bar{P}_{\ts{a}}$ and $\bar{P}_{\ts{p}}$ denote the means of $P_{\ts{a}}$ and $P_{\ts{p}}$, respectively. Here, it can be seen that power consumption stays approximately constant with respect to actuation frequency. In air, $\bar{P}_{\ts{a}}$ and $\bar{P}_{\ts{p}}$ are on the order of \mbox{$40$\,mW} and \mbox{$500$\,mW}, for a PWM excitation of $1\,\ts{Hz}$; in water, $\bar{P}_{\ts{a}}$ and $\bar{P}_{\ts{p}}$ are on the order of \mbox{$0.9$\,W} and \mbox{$10.7$\,W}, for a PWM excitation of $1\,\ts{Hz}$, which corresponds to a \mbox{$2150\,\%$} increase in average power consumption with respect to the \mbox{actuation-in-air} case. This significant increase in power is explained by the additional energy required to Joule heat the SMA wires of the actuator under water due to the high \mbox{heat-transfer} coefficient of this medium. This phenomenon can be observed in the actuator responses shown in~\mbox{Fig.\,\ref{FIG02}(f)}. In the \mbox{actuation-in-air} case, the instantaneous power, $P(t)$, remains below $600\,\ts{mW}$ while in the \mbox{actuation-in-water} case, the instantaneous power, $P(t)$, exceeds $10\,\ts{W}$ in three instances during the $3$ seconds shown in the plot. Furthermore, the phase transition from \textit{austenite} to \textit{martensite} of the SMA material is drastically faster in water than in air due the significantly higher cooling rate in the former medium than in the latter. Video footage showing the actuator functioning in both air and water can be seen in the accompanying supplementary movie.

\section{Novel \mbox{Low-Power} \mbox{SMA-Based} Actuation for Underwater Propulsion}
\label{SECTION03}
To solve the problem of power for \mbox{SMA-based} microactuation in water, we propose the conceptual design depicted at the bottom of~\mbox{Fig.\,\ref{FIG02}(a)}. The key element of this approach is an insulating air chamber made of Kapton that surrounds the SMA wires that drive the actuator, which locally reduces the \mbox{heat-transfer} coefficient of the system. Basic thermal analyses indicate that this design would drastically lower the power needed to efficiently generate \mbox{SMA-based} microactuation in water, thus enabling the development of submersible microswimmers. Using the same setup depicted in~\mbox{Fig.\,\ref{FIG02}(b)}, we determined that a \mbox{first-generation} \mbox{$13$-mg} prototype of this design can operate at a frequency of $1\,\ts{Hz}$ consuming only an average power of \mbox{$150\,\ts{mW}$} with a peak of $1.6\,\ts{W}$. This approach clearly shows a path towards autonomous underwater microrobotics.

\bibliographystyle{IEEEtran}
\bibliography{references}
\end{document}